\documentclass[pdflatex,sn-mathphys-num]{sn-jnl}

\usepackage{fix-cm}%
\usepackage{graphicx}%
\usepackage{multirow}%
\usepackage{amsmath,amssymb,amsfonts}%
\usepackage{amsthm}%
\usepackage[scr]{rsfso}%
\usepackage[title]{appendix}%
\usepackage{xcolor}%
\usepackage{textcomp}%
\usepackage{manyfoot}%
\usepackage{booktabs}%
\usepackage{algorithm}%
\usepackage{algorithmicx}%
\usepackage{algpseudocode}%
\usepackage{listings}%
\usepackage{tabularx}
\usepackage{color}
\usepackage{hyperref}

\theoremstyle{plain}%
\theoremstyle{remark}%

\theoremstyle{definition}%

\begin{document}

\title[Article Title]{FwNet-ECA: A Classification Model Enhancing Window Attention with Global Receptive Fields via Fourier Filtering Operations}


\author[1]{\fnm{Shengtian} \sur{Mian}}\email{22023120124@cueb.edu.cn}

\author[2]{\fnm{Ya} \sur{Wang}}\email{wangyachn@pku.edu.cn}

\author[1]{\fnm{Nannan} \sur{Gu}}\email{gu\_nannan@126.com}

\author[1]{\fnm{Yuping} \sur{Wang}}\email{wangyuping@cueb.edu.cn}

\author*[1]{\fnm{Xiaoqing} \sur{Li}}\email{xqli@cueb.edu.cn}

\affil[1]{\orgdiv{School of Statistics}, \orgname{Capital University of Economics and Business}, \orgaddress{\city{Beijing}, \postcode{100070},  \country{China}}}

\affil[2]{\orgdiv{School of Mathematical Sciences}, \orgname{Peking University}, \orgaddress{\city{Beijing}, \postcode{100871},  \country{China}}}


\abstract{Windowed attention mechanisms were introduced to mitigate the issue of excessive computation inherent in global attention mechanisms. In this paper, we present FwNet-ECA, a novel method that utilizes Fourier transforms paired with learnable weight matrices to enhance the spectral features of images. This method establishes a global receptive field through Filter Enhancement and avoids the use of moving window attention. Additionally, we incorporate the Efficient Channel Attention (ECA) module to improve communication between different channels. Instead of relying on physically shifted windows, our approach leverages frequency domain enhancement to implicitly bridge information across spatial regions. We validate our model on the iCartoonFace dataset and conduct downstream tasks on ImageNet, demonstrating that our model achieves lower parameter counts and computational overheads compared to shifted window approaches, while maintaining competitive accuracy. Furthermore, our visualization operations clearly demonstrated that the Filter Enhancement technique achieves greater effectiveness in the model's shallow layers, where feature maps are relatively larger. This work offers a more efficient and effective alternative for leveraging attention mechanisms in visual processing tasks, alleviating the challenges associated with windowed attention models. Code is available at \url{https://github.com/qingxiaoli/FwNet-ECA}}




\maketitle

\section{Introduction}

The Transformer architecture, firstly proposed in the realm of Natural Language Processing (NLP), has in recent years been adopted into the CV(computer vision), where it has exhibited superior performance over Convolutional Neural Networks (CNNs) on large-scale datasets. Vision Transformer models leverage Self Attention layers to establish dependencies among all pixels in an image, thereby capturing extensive spatial information.

However, the pursuit of capturing global spatial relationships by Self Attention models incurs a quadratic increase in computational complexity with respect to image resolution. A common approach to mitigate this involves encoding neighboring pixels into a single token. Despite this, the model still demands considerable parameters and computation, with limitation particularly evident in tasks such as image segmentation and object detection. Liu et al. \cite{liu2021swin}addressed this by partitioning images into smaller pixel regions and employing windowed attention to reduce computational load. To overcome the restricted receptive field of local attention, they further proposed shifted window attention for cross-window information exchange. However, this method only enhances the connections between partial windows and lacks a global receptive field. Moreover, the implementation of shifted windows involves complex masking operations and tensor shape transformations, making it relatively complicated. We aim to achieve a similar effect in a simpler way.

In this work, we use filter enhancement operations to establish a global receptive field for window attention, thereby avoiding the use of shifted window attention. The purpose of using filtering operations is to enhance or suppress certain frequencies, allowing the window attention to more clearly identify where to focus its attention. Following each Filtering Enhancement, we append adjacent channel attention to bolster inter-channel interactions, termed FwNet-ECA. Unlike the moving window attention, the filter enhancement layer directly obtains a global receptive field, and the computation is simple, requiring only Fourier transforms and element-wise matrix multiplication. This is easy to implement based on the PyTorch library. Leveraging the Fast Fourier Transform (FFT) in two dimensions, the computational complexity of the Filter Enhancement layer is reduced to O(NlogN).

Our model's efficacy is demonstrated  on the iCartoonFace dataset and downstream tasks of ImageNet, where we achieve competitive performance with significantly fewer parameters and reduced computational cost when benchmarked against prominent models including ViT\cite{dosovitskiy2020image}, ResMLP\cite{touvron2022resmlp}, and ResNet\cite{he-2016}.

We further substantiate the effectiveness of the filter enhancement operation, especially in the shallow layers of models where feature maps are relatively large, through a series of visualization experiments. This finding demonstrates that employing filter enhancement techniques in the initial stages of a model can significantly improve the quality of feature extraction, thereby enhancing overall model performance. This effect is particularly beneficial for tasks that require capturing more macroscopic and global information from input images.

\section{Related Works}
\begin{figure}[htbp]
    \centering
    \includegraphics[width=1\linewidth]{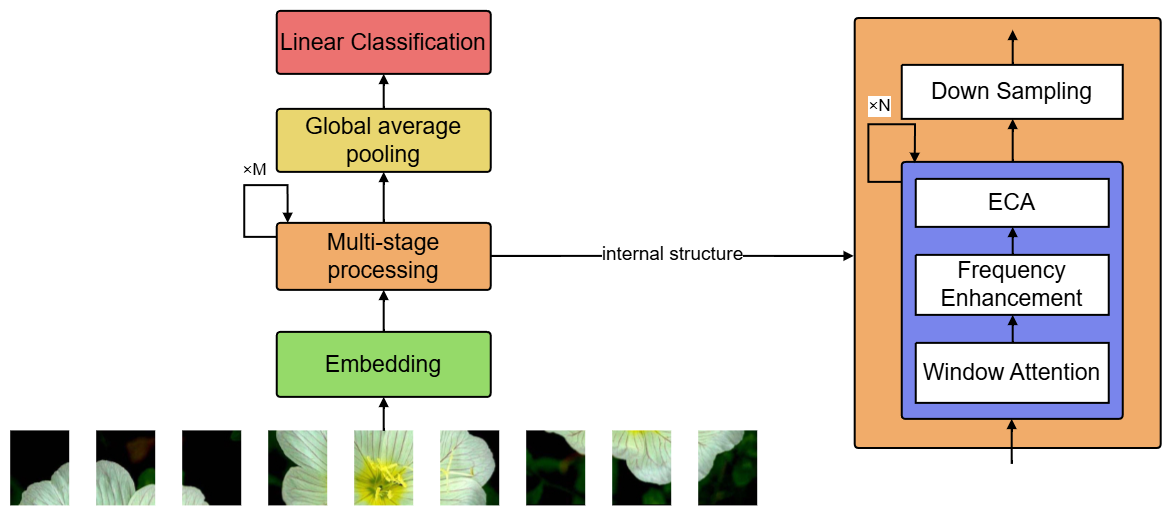}
    \caption{Overview of FwNet-ECA Architecture}
    \label{fig:1}
\end{figure}

CNNs have been instrumental in advancing computer vision, offering a robust framework for image recognition\cite{lecun2015deep}. Their hierarchical feature extraction capabilities, which are designed to mimic the human visual cortex, have proven effective across a wide range of applications\cite{serre2007feedforward}. Key developments in CNN architectures\cite{krizhevsky2012imagenet}\cite{simonyan2014very}\cite{szegedy2015going}\cite{he-2016}, have progressively improved performance in image classification, detection, and segmentation tasks. However, CNNs' reliance on convolutional layers for feature extraction inherently limits their capacity to model long-range dependencies efficiently, a critical aspect for tasks requiring global context understanding\cite{long2015fully}. 

The advent of Transformers in natural language processing has catalyzed a paradigm shift in computer vision, leading to the exploration of ViT\cite{vaswani2017attention}. Pioneered by Dosovitskiy et al. \cite{dosovitskiy2020image}, ViT treat images as sequences of patches, enabling more flexible modeling of global dependencies without the need for convolutional operations. This approach achieved impressive results in image classification but was inherently  constrained by high computational requirements and large model sizes\cite{touvron2021training}. Subsequent improvements aimed to tackle these issues through architectural refinements, often reintroducing some level of inductive bias. A notable example is the Swin Transformer\cite{liu2021swin}, which curtails computational complexity via localized attention mechanisms. To facilitate interaction between different windows, it introduces shifted window attention, which is a very clever method but is complex in operation. In this work, we aim to propose a simpler and more intuitive method.

In the visual domain, the Fourier transform, particularly the Discrete Fourier Transform (DFT), appears in many convolutional neural networks. With CNNs demonstrating substantial advancements in visual recognition and analysis, various strategies have emerged to optimize and innovate for specific visual tasks. One such method involves exploiting frequency domain representations through DFT, thereby enhancing task performance by leveraging frequency features\cite{lee2018single}\cite{yang2020fda}. Another concurrent approach leverages the convolution theorem in conjunction with FFT to expedite CNN computations\cite{li2020falcon}. Within Transformer architectures, networks like FNet\cite{lee2021fnet} and GFNet\cite{rao2021global} have replaced Self Attention mechanisms with filtering operations for both CV and NLP tasks.

Regarding channel attention, SENet\cite{hu2018squeeze} initiated the concept of channel-wise attention, enabling networks to dynamically adjust feature channel weights through its SE(Squeeze-and-Excitation) module, significantly boosting model performance. Subsequent innovations, such as ECA-Net\cite{wang2020eca} and CBAM\cite{woo2018cbam}, not only improved computational efficiency but also fortified feature processing capabilities by integrating spatial attention. In this work, we added an ECA(Efficient Channel Attention) module to the main structure of the model to construct inter channel attention.

Notably, in contrast to similar efforts like FNet and GFNet, FwNet-ECA distinguishes itself in several key aspects:
\begin{itemize}
\item \textbf{Retention of Self-Attention:} Unlike FNet and GFNet, which abandon Self Attention entirely, FwNet-ECA integrates partial Self Attention mechanisms, leveraging their proven effectiveness in both NLP and CV.
\item 	\textbf{Channel Interaction:} FwNet-ECA addresses the shortfall of FNet and GFNet by explicitly considering inter-channel relationships through mechanisms like local channel attention.
\item 	\textbf{Domain Adaptation:} Unlike FNet, which is tailored for NLP tasks, FwNet-ECA is purpose-built for CV applications, demonstrating versatility and domain-specific optimization.
\item 	\textbf{Different purposes:} Although the method is similar to GFNet, the purpose of constructing FwNet-ECA is to find a simple and efficient method to replace the more complex moving window attention.
\end{itemize}

\section{Method}

\subsection{Discrete Fourier transform}

Our approach is grounded in the Two-Dimensional Discrete Fourier Transform (2D DFT), though for initial comprehension, let us consider the One-Dimensional Discrete Fourier Transform (1D DFT). For a one-dimensional sequence of length \(N\), it can be decomposed into a sum of \(N\) frequency components. This transformation is mathematically encapsulated by \eqref{DFT}.
\begin{equation}
    F[k]=\sum_{n=0}^{N-1}f[n]\cdot e^{-j\frac{2\pi}{N}k n}\label{DFT}
\end{equation}

In the context of the Discrete Fourier Transform (DFT): 
\begin{itemize}
\item \(F[k]\) represent the complex amplitude of the \(k\) th frequency component, where \(k\)  ranges from  \(0\) to \(N-1\). 
\item \(f[n]\) denote the \(n\) th sample of the original signal, with \(n\) ranging from  \(0\) to \(N-1\).
\item \(N\) is the total number of samples in the signal and also the number of frequency components. 
\item 	\(e^{-j\frac{2\pi}{N}k n}\) is the complex exponential term, with \(j\)  being the imaginary unit and \(\frac{2\pi}{N}\) being the angular frequency increment.
\end{itemize}

Owing to the conjugate symmetry property of Fourier Transform, the complex plane is \(F[N-k]=F^{*}[k]\), then the number of frequency domain data matrices and weight matrices decreases from \(F [N]\) to \(F [N/2]\). Only half of the data is necessary to retain all the information about the image, thereby inherently reducing parameter counts and computational demands to a certain extent.

The computational complexity of the DFT is \(O(N^{2})\) ; However, the FFT exploits the symmetry and periodicity inherent in Equation \eqref{DFT} to optimize this, reducing the complexity dramatically to \(O(Nlog(N))\). 

Moreover, the reversibility of the Fourier Transform ensures bidirectional information transfer between the frequency domain and the spatial domain, facilitating flexible manipulation and analysis of data representations. This depends on the \eqref{IDFT}.
\begin{equation}
f[n]={\frac{1}{N}}\sum_{k=0}^{N-1}F[k]e^{j(2\pi/N)k n}\label{IDFT}
\end{equation}

Extending to two dimensions, Fourier Transform (FT) conceives fitting a complex surface with a series of periodic waveforms, as formulated following \eqref{2D-DFT}.
\begin{equation}
    F(u,v)=\sum_{m=0}^{M-1}\sum_{n=0}^{N-1}f(m,n)\cdot e^{-j2\pi\left(\frac{u m}{M}+\frac{v n}{N}\right)}\label{2D-DFT}
\end{equation}

\begin{equation}
    f(m,n)=\frac{1}{M N}\sum_{m=0}^{M-1}\sum_{n=0}^{N-1}{F}(u,v)\cdot e^{j2\pi\left(\frac{u m}{M}+\frac{v n}{N}\right)}\label{2D-IDFT}
\end{equation}

\begin{table}[htbp]
\centering
\caption{Comparison of Computational Complexity: Filtering Enhancement Layer vs. Other Methods}
\label{tab1}
\begin{tabular}{@{}ll@{}} 
\toprule
\textbf{Method}                                  & \textbf{Complexity}   \\ \midrule
Multiple head self attention                     & $4HWC^{2}+2(HW)^{2}C$ \\
Window-Multiple head self attention              & $4HWC^{2}+2M^{2}HWC$  \\
\textbf{Filter Enhancement}& $2HWC[\log HW]+HWC$ \\ \bottomrule
\end{tabular}
\begin{tablenotes}
\item $^{\mathrm{1}}$ $H$ is the height, $W$ is the width, $C$ is the dimension of the token, and $M$ is the size of the window.
\end{tablenotes}
\end{table}

This transformation decomposes the original image into a sum of sinusoidal patterns, each characterized by a specific spatial frequency. Low frequencies correspond to slow variations in the image (like gradual changes in brightness), while high frequencies represent sharp edges and fine details. The two-dimensional discrete Fourier Transform retains the reversibility \eqref{2D-IDFT} and conjugation \(F[M-u.N-v]=F^{*}[u,v]\),  It is worth mentioning that the conjugation in the spectrum of two-dimensional Fourier Transform can be used in high or wide dimensions, that is,  \(F(u,v)\) to \(F(u,v/2)\) or \(F(u,v)\) to \(F(u/2,v)\). It can also be accelerated by two-dimensional fast Fourier transform, thus reducing the amount of calculation from \(O(MN^{2})\) to \(O(MNlog(MN))\). We compared the computational complexity of different methods and the results are shown in Table \ref{tab1}.

\subsection{FwNet-ECA}

\begin{figure}[htbp]
    \centering
    \includegraphics[width=0.5\linewidth]{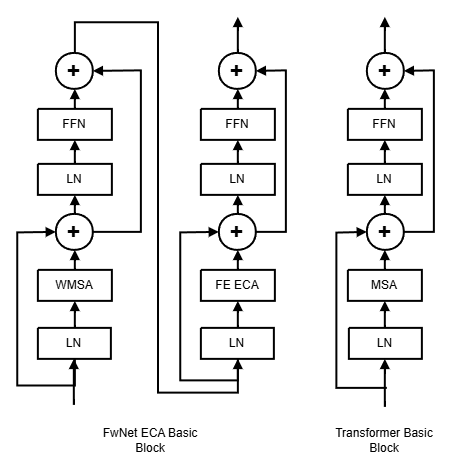}
    \caption{Comparison of FwNet-ECA Basic Module vs. Traditional Transformer Basic Module}
    \label{fig:2}
\end{figure}

FwNet-ECA, an architecture that builds upon the evolution of Transformer models, is designed with a dual-core structure in each of its fundamental modules. These cores consist of Windowed Self Attention and Filter Enhancement, collectively contributing to a sophisticated processing pipeline. Following each attention and enhancement layer, a Feed-Forward Network (FFN) is integrated to distill semantic information from the processed features, as depicted in the conceptual schema of Module Structure, Fig.~\ref{fig:2}. In the figure, WMSA refers to Window Multi Head Self Attention, and FE ECA refers to Filter Enhancement and Efficient Channel Attention, MSA refers to Multi Head Self Attention. Compared with the original transformer architecture, our model connects a Filter Enhancement layer after Window Self Attention to realize cross window communication. 

The Filter Enhancement Layer operates by first applying a 2D DFT to the image data, translating spatial information into the frequency domain. The weighting matrix \(W\) then selectively reweights various frequency components, and then the inverse DFT restores the information back to the spatial domain to ensure compatibility with the subsequent FFN\cite{vaswani2017attention} and the window attention layer. The mathematical representation underlying this transformation process is outlined by Equation $F=H^{-1}[W\odot H[F]]$, details are shown in formula \eqref{FE}, which encapsulates the essence of information flow from spatial to frequency domains and back.

\begin{equation}
\begin{aligned}
F^{r}(u,v) &= \sum_{m=0}^{M-1}\sum_{n=0}^{N-1}f^{r}(m,n)\cdot e^{-j2\pi\left(\frac{u m}{M}+\frac{v n}{N}\right)},\\
{\tilde{F^{r}}}(u,v) &= W(u,v)\cdot F^{r}(u,v),\\
{\tilde f^{r}}(m,n) &= \frac{1}{M N}\sum_{m=0}^{M-1}\sum_{n=0}^{N-1}{\tilde{F^{r}}}(u,v)\cdot e^{j2\pi\left(\frac{u m}{M}+\frac{v n}{N}\right)}.
\end{aligned}
\label{FE}
\end{equation}

Window Self Attention calculation is performed before each Filter Enhancement layer, and the window Self Attention layer has not undergone any special processing. The process of this layer is shown in formula \eqref{WA}.

\begin{equation}
\begin{aligned}
{\hat{f^{r}}}(m,n) &= \mathrm{WMSA}(LN({f^{r-1}}(m,n)))+{f^{r-1}}(m,n),\\
{f^{r}}(m,n) &= \mathrm{FFN}(LN({\hat{f^{r}}}(m,n)))+{\hat{f^{r}}}(m,n)).
\end{aligned}
\label{WA}
\end{equation}

The Local Channel Attention module fosters inter-channel communication by focusing on attention between neighboring channels. This is achieved by globally averaging each channel, applying a 1x3 convolution filter and using the sigmoid activation function to generate weights for channel-wise multiplication. This mechanism enhances cross-channel dependencies, enriching feature representations. Mathematical details are shown in the formula \eqref{ECA}.

Inspired by Swin Transformer\cite{liu2021swin}, FwNet-ECA also collapses 4x4 pixel blocks into tokens, employs relative positional encoding within window-based attention, and incorporates Patch Merging for downsampling at the conclusion of each stage. The architecture of the model is shown in Fig.~\ref{fig:1}. It should be noted that at the end of the last stage, the down-sampling is not performed, but the classification is performed after the global average pooling.

\begin{equation}
\begin{aligned}
g^{r}_{c} &= \mathrm{GlobalAvgPool}({\tilde f^{r}}(m,n)), \\
h^{r}_{c} &= \mathrm{Conv1d}(g^{r}_{c}, \mathrm{kernel\;size}=3), \\
w^{r}_{c} &= \sigma(h^{r}_{c}), \\
\tilde{f}^{r}_{c}(m,n) &= w^{r}_{c} \odot {\tilde f^{r}}(m,n).
\end{aligned}
\label{ECA}
\end{equation}

Notably, FwNet-ECA achieves competitive or superior performance compared to mainstream Transformer models with significantly fewer parameters and reduced computational overhead, all without incorporating prior knowledge. This design choice potentially endows the model with heightened learning capabilities.

\section{Experiment}

\begin{table}[htbp]
\centering
\caption{Composition of Datasets Used in the Study}
\label{tab2}
\begin{tabularx}{\textwidth}{l *{4}{>{\centering\arraybackslash}X}} 
\toprule
\textbf{Dataset}       & \textbf{Total number of images} & \textbf{Train number of images} & \textbf{Val number of images} & \textbf{Number of categories} \\
\midrule
iCartoonFace\cite{zheng2020cartoon}  & 389678                 & 353029                 & 36649                & 5013                 \\
CIFAR100\cite{krizhevsky2009learning}      & 60000                  & 50000                  & 10000                & 100                  \\
Flowers\cite{nilsback2008automated}       & 8189                   & 2040                   & 6149                 & 102                  \\
Stanford Cars\cite{krause20133d}          & 16185                  & 8144                   & 8041                 & 196                  \\
\bottomrule
\end{tabularx}
\end{table}

We conducted comparative evaluations with various models on datasets including icartoonface, CIFAR100, Stanford Cars-196, and Flowers-102 to validate the effectiveness of FwNet-ECA. Detailed descriptions of these datasets are provided in Table \ref{tab2}. The icartoonface data set is randomly divided into the training set and the verification set at a ratio of 9:1.

Across all model training processes, we employed a series of data augmentation techniques to improve the generalization ability of our models. Specifically, these included resizing images to a fixed size $(224, 224)$, applying a \texttt{RandomResizedCrop} with scale ranging from $0.15$ to $1.0$ and aspect ratio between $3/4$ and $4/3$, using \texttt{RandomHorizontalFlip} for horizontal flipping, and utilizing \texttt{ColorJitter} to randomly adjust the brightness, contrast, and saturation within a range of $0.4$. Additionally, images were converted into tensors and normalized using mean values of $[0.485, 0.456, 0.406]$ and standard deviation values of $[0.229, 0.224, 0.225]$. Our models were initialized with pre-trained weights from ImageNet1k. We compared the computational cost, parameter counts, and Top-1 accuracy of different models.

\subsection{Icartoonface Classification and Ablation Experiment}

\begin{table}[htbp]
\centering
\caption{Comparison of Parameter Counts, Computational Costs, and Top-1 Accuracy Across Models on the iCartoonFace Dataset with Ablation Experiment}
\label{tab3}
\begin{tabular}{lccc} 
\toprule
\textbf{Network}                & \textbf{Params(M)} & \textbf{FLOPs(G)} & \textbf{Top-1 Acc(\%)} \\
\midrule
Densenet-169-F-ArcFace\cite{zheng2020cartoon} & 14.3      & 3.4      & 84            \\
Gfnet-S\cite{rao2021global}                & 25        & 4.5      & 95.8          \\
Swin-T\cite{liu2021swin}                 & 28.3      & 4.4      & 95.9          \\
win                                    & 23.9      & 3.7      & 95.3          \\
FwNet-T                                 & 24.6      & 3.7      & 95.9          \\
FwNet-SE-T                              & 25.7      & 3.7      & 96.1          \\
FwNet-ECA-T                             & 24.6      & 3.7      & \textbf{96.1} \\
\bottomrule
\end{tabular}
\begin{tablenotes}
\item $^{\mathrm{a}}$ The win model does not perform Filter Enhancement operations.
\end{tablenotes}
\end{table}

Table \ref{tab3} presents the results of various models under the same number of parameter updates (50 epochs). When compared to models with similar parameter sizes, such as Swin-T and GFNet-S, our model achieved comparable or slightly better results, highlighting that FwNet-ECA matches or outperforms these models in terms of accuracy while requiring significantly less computational resources and having fewer parameters. It is worth noting that GFNet also employs Fourier transforms, similar to our approach; however, despite its innovative exclusion of self-attention, it does not surpass the performance of self-attention models. Our method retains the benefits of both approaches while reducing the computational load and maintaining the effectiveness. Additionally, when compared to the intricately designed DenseNet, our model outperformed it by  14$\%$ in accuracy while maintaining a similar computational load.

Ablation studies were also performed on the Filter Enhancement layer and the channel attention module. The results show that the performance of FwNet with the Filter Enhancement layer is better than that of the baseline model (win) without the Filter Enhancement operation. Regarding channel attention, models utilizing both the SE module and the ECA module achieved equal or improved accuracy over the original FwNet; however, given the higher parameter count of the SE module compared to ECA, the selection of ECA is justified for better efficiency. 

\subsection{Imagenet downstream tasks}

\begin{table}[htbp]
\centering
\caption{Comparison of Parameter Counts, Computational Costs, and Top-1 Accuracy(\%) Across Models in Downstream Tasks}
\label{tab4}
\begin{tabular}{lcccccc} 
\toprule
\textbf{Network}       & \textbf{Params(M)} & \textbf{FLOPs(G)} & \textbf{CIFAR100} & \textbf{Flowers} & \textbf{Cars} \\
\midrule
ResNet-50\cite{he-2016}     & 26        & 4.1      & -             & 96.2          & 90            \\
ResNet-50-SAM\cite{chen2021vision} & 26        & 4.1      & 85.2          & -             & -             \\
ResMLP-24\cite{touvron2022resmlp}     & 30        & 6        & \textbf{89.5} & 97.9          & 89.5          \\
Mixer-S/16\cite{chen2021vision}    & 18.5      & 3.78     & 77.9          & 83.3          & -             \\
Mixer-B/16\cite{chen2021vision}    & 59.9      & 12.6     & 80            & 82.8          & -             \\
ViT-B/16\cite{dosovitskiy2020image}      & 86        & 55.4     & 87.1          & 89.5          & -             \\
ViT-L/16\cite{dosovitskiy2020image}      & 307       & 190.7    & 86.4          & 89.7          & -             \\
FwNet-ECA-T   & 24.6      & 3.7      & 85.4          & 97.7          & \textbf{90.4} \\
FwNet-ECA-S   & 33.8      & 5.4      & -             & \textbf{98.2} & -             \\
\bottomrule
\end{tabular}
\end{table}

To assess the model's transfer learning capability, we evaluated it on downstream datasets of ImageNet, specifically CIFAR-100, Flowers-102, and Stanford Cars. Following the icartoonface training protocol, no additional data augmentation was applied, and models were fine-tuned after initialization with pre-training weights. As shown in Table \ref{tab4}, compared to ResMLP, our model performs slightly worse on CIFAR-100. This is attributed to the fact that ResMLP employs extensive data augmentation techniques during training, whereas our model only implements simple random cropping and color jittering. Furthermore, ResMLP leverages Knowledge Distillation to enhance the model’s accuracy, allowing it to learn more categories with reduced computational overhead. Consequently, ResMLP outperforms our model in terms of accuracy on the CIFAR-100 dataset, which exhibits significant inter-class differences. On the other hand, our model surpasses ResMLP on the Flowers and Cars datasets, supporting our hypothesis that our model is naturally suited for fine-grained image classification tasks. Unfortunately, due to computational resource constraints, we were unable to train a computationally intensive teacher model for Knowledge Distillation. 

\begin{table}[htbp]
\centering
\caption{Comparison of Model Settings with Different Sizes}
\label{tab5}
\begin{tabular}{lcccc} 
\toprule
       & \textbf{FwNet-T}   & \textbf{FwNet-S}    & \textbf{FwNet-B}    & \\
\midrule
stage1 & Block $\times$ 2 & Block $\times$ 2  & Block $\times$ 2  & \\
stage2 & Block $\times$ 2 & Block $\times$ 2  & Block $\times$ 2  & \\
stage3 & Block $\times$ 6 & Block $\times$ 12 & Block $\times$ 18 & \\
stage4 & Block $\times$ 2 & Block $\times$ 2  & Block $\times$ 2  & \\
params(M) & 24.6      & 33.8       & 42.8       & \\
Flops(G)  & 3.7       & 5.4        & 7.1        & \\
\bottomrule
\end{tabular}
\end{table}

\begin{figure}[htbp]
    \centering
    \includegraphics[width=0.75\linewidth]{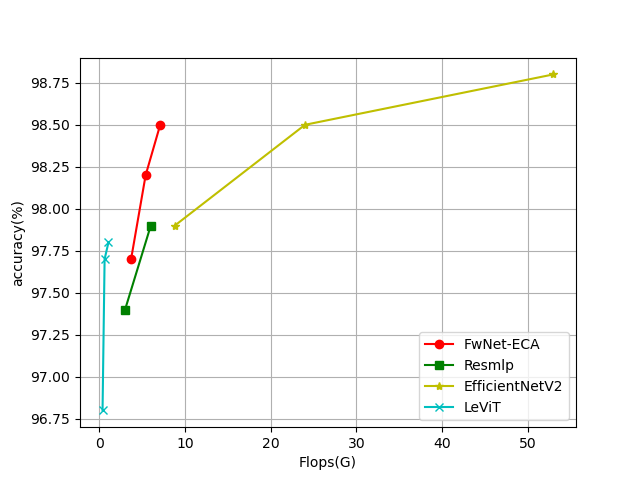}
    \caption{Comparison of Model Accuracy on the Flowers-102 Dataset Across Different Computational Costs}
    \label{fig:3}
\end{figure}

We also compared the top-1 accuracy of different models at various scales. Large-scale model training often demands substantial computational resources, and due to these constraints, we conducted our comparisons on the Flowers-102 dataset. The specific model configuration details are provided in Table \ref{tab5}. The results are illustrated in Fig.~\ref{fig:3}. Our model outperforms other carefully designed models, such as EfficientNet\cite{tan2021efficientnetv2}, at low computational costs. However, at even lower computational levels, LeViT\cite{graham2021levit} demonstrates superior performance. To scale up our FwNet-ECA model, we propose increasing the token dimension and stacking more Basic Modules. Based on the designs of other models, we believe that stacking Basic Modules in the second or third layer is more beneficial.

\subsection{Visualization}

To more clearly demonstrate how our filter enhancement boosts the establishment of a global receptive field to assist window attention, we visualized the responses of images from the filter enhancement modules and shifted window attention modules at different stages in both FwNet-ECA and Swin Transformer. For this visualization, we employed three types of Class Activation Mapping (CAM) methods: GradCAM\cite{selvaraju2017grad}, XGradCAM\cite{jacobgilpytorchcam}, LayerCAM\cite{jiang2021layercam}, and the outputs of each method were averaged to obtain the Integrated CAM. The rationale behind choosing these methods is that while GradCAM is suitable for rapid implementation and broadly applicable, it may underperform in complex scenarios. XGradCAM, on the other hand, improves accuracy and interpretability through normalization, making it ideal for tasks requiring high-precision activation maps. LayerCAM, which integrates feature maps from multiple layers, is particularly apt for complex scenes and multi-scale analysis. Concerning model selection, we utilized a tiny-sized model with pre-training conducted on the iCartoonFace dataset. As shown in \ref{tab5}, stages 1 and 4 each contain only one filter enhancement or shifted window attention block, whereas in stage 3, we selected the middle module for examination.

\begin{figure}[htbp]
    \centering
    \includegraphics[width=1\linewidth]{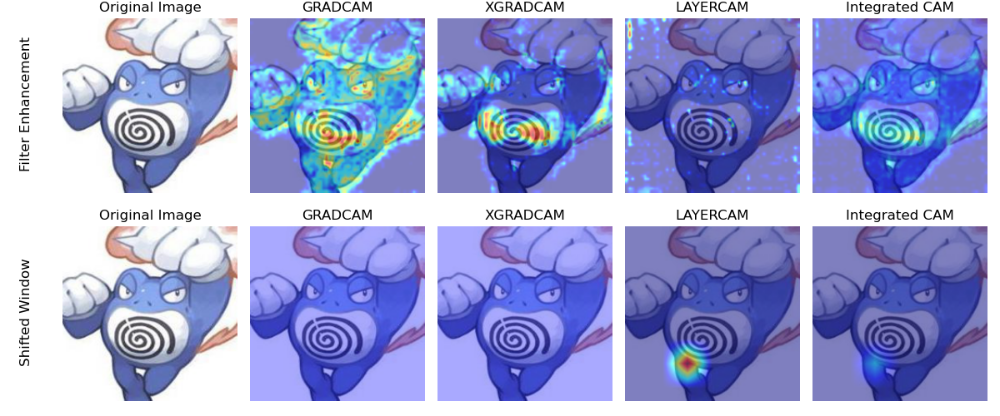}
    \caption{Visual Comparison of Filter Enhancement Operations vs. Shifted Window Attention in the Initial Stage of the Model}
    \label{layer0}
\end{figure}

We chose this image for visualization because it possesses high resolution and distinct features, such as unique lines and an exaggerated artistic style, making it ideal for observation through visualization techniques. As shown in GradCAM of Fig. \ref{layer0}, our frequency enhancement operation in stage 1 demonstrates a grasp on the overall outline of the character and key features like the curve of the character's abdomen. XGradCAM highlights the main focus of the model by effectively reducing noise; however, our model does not show significant responses in images generated by LayerCAM. This is likely due to the relatively uniform gradients in the shallow layers of our model, whereas LayerCAM calculates gradient information for each pixel, aiming to capture finer-grained local responses. This also explains why the visualization of shifted window attention shows no response in GradCAM and XGradCAM but reveals a small area of interest in LayerCAM. Although the gradients of the shifted window attention are minimal, there is a notable change in a localized region at the lower left part of the character's abdomen in the image.

\begin{figure}[htbp]
    \centering
    \includegraphics[width=1\linewidth]{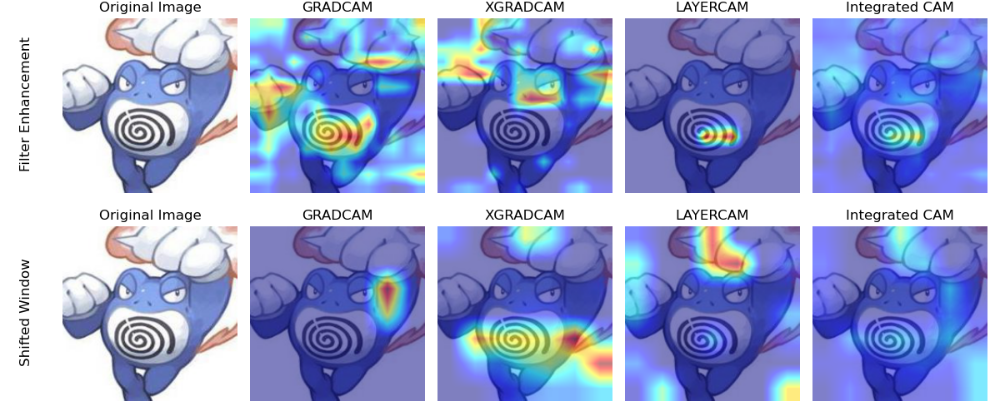}
    \caption{Visual Comparison of Filter Enhancement Operations vs. Shifted Window Attention in the Middle Stage of the Model}
    \label{layer2}
\end{figure}
\begin{figure}[htbp]
    \centering
    \includegraphics[width=1\linewidth]{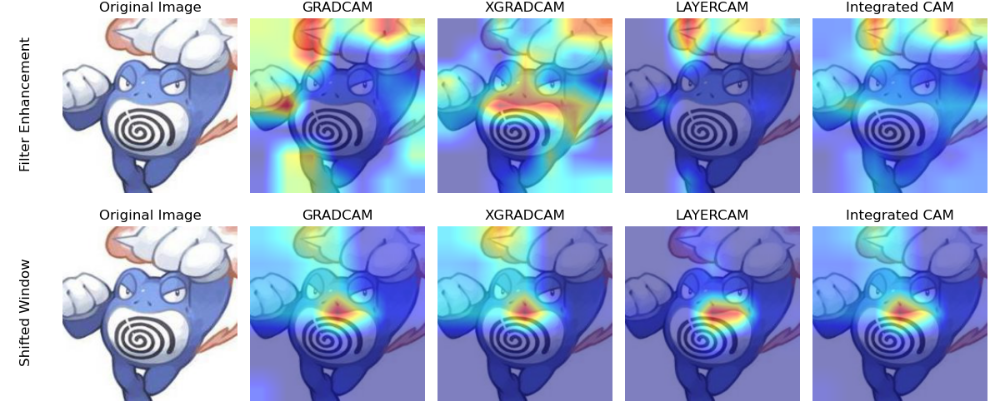}
    \caption{Visual Comparison of Filter Enhancement Operations vs. Shifted Window Attention in the Last Stage of the Model}
    \label{layer3}
\end{figure}

As we've conducted visualizations in the intermediate and final stages of the model, as shown in Fig. \ref{layer2} \ref{layer3} respectively, it's evident that the filter enhancement responses generally cover a larger area and respond at more positions across the image compared to the shifted window approach. The response locations of the shifted window method show a clear blocky distribution, which is especially noticeable in the final stage. In these stages, it can be seen that the shifted window responses are concentrated in the central region and a square area in the upper left of the image.

This observation suggests that while filter enhancement tends to activate over wider areas and potentially capture broader or more diverse features within an image, the shifted window attention mechanism focuses its responses more narrowly, possibly honing in on specific features or areas of interest with higher precision. This distinction highlights the different strategies these methods employ for feature extraction and their implications on the model's understanding of the input images.

\subsection{Analysis}

\begin{figure}
    \centering
    \includegraphics[width=0.75\linewidth]{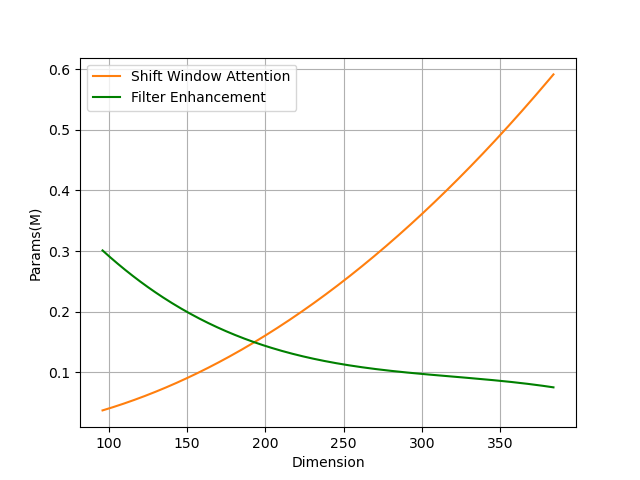}
    \caption{Parameter Quantity Variation Across Different Dimensions}
    \label{fig:4}
\end{figure}

\begin{figure}
    \centering
    \includegraphics[width=0.75\linewidth]{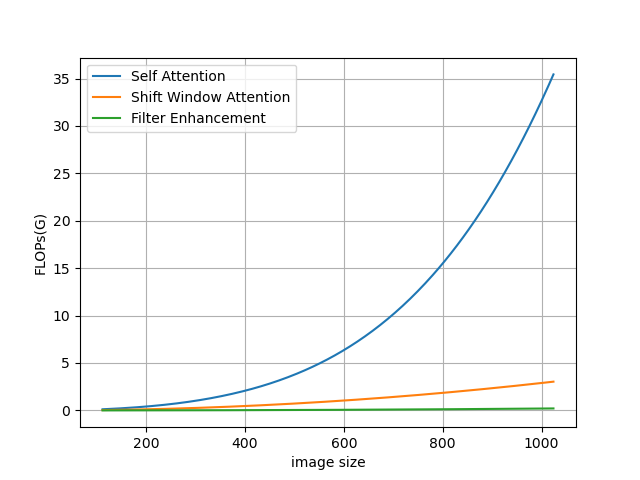}
    \caption{Comparison of FLOPs Across Different Image Resolutions}
    \label{fig:5}
\end{figure}

\begin{figure}
    \centering
    \includegraphics[width=0.75\linewidth]{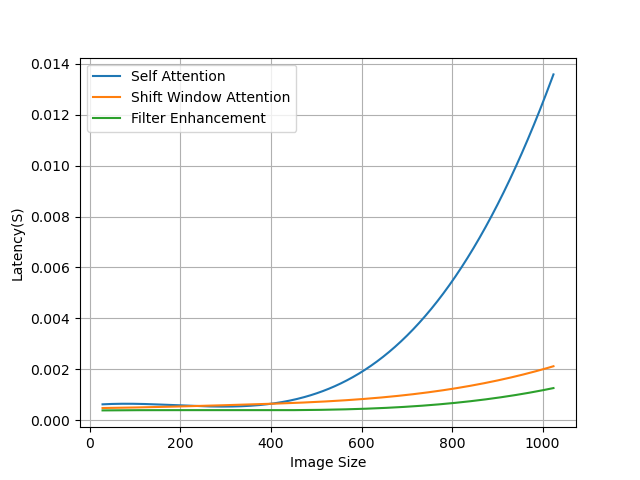}
    \caption{Output Latency Variations Among Different Methods}
    \label{fig:6}
\end{figure}

In the following analysis, it should be noted that for aesthetic reasons, the curves were smoothed when drawing Fig.\ref{fig:4}~\ref{fig:5}~\ref{fig:6}. To analyze the changes in the number of parameters and computational complexity of filter enhancement, self-attention, and windowed self-attention with respect to the feature map size and token dimensions. We conducted a detailed analysis of the Filter Enhancement layer within a hierarchical model structure, using the same patch size and initial token dimension for embedding operations. In a hierarchical structure, the token dimension is doubled after each stage, and the height and width of the feature map are halved, leading to a rapid increase in dimensionality. The change in parameter count with increasing token dimensions is illustrated in Fig.~\ref{fig:4}. Since the windowed self-attention does not change the number of parameters compared to self-attention, we have only plotted the curve for the variation of the number of parameters for self-attention. Initially, the parameter count for Filter Enhancement exceeds that of Self Attention; however, as the token dimension increases, the parameter count for Self Attention grows exponentially, whereas the parameter count for Filter Enhancement decreases. This suggests that Filter Enhancement is more sensitive to image resolution or token quantity, while Self Attention is more sensitive to token dimension.

Furthermore, we compared the global attention (ViT), windowed attention (Swin), and Filter Enhancement methods in terms of computational cost and output latency for the smallest models described in their respective papers. The global attention method uses a patch size of 16x16 and a token dimension of 768. The windowed attention method uses a patch size of 4x4 and a token dimension of 96. Similarly, the Filter Enhancement method uses a patch size of 4x4 and a token dimension of 96. As shown in Fig.~\ref{fig:5}, windowed attention already demonstrates a significant reduction in computational cost compared to global attention, yet it still remains higher than Filter Enhancement. Fig.~\ref{fig:6} presents the output latency for different resolutions, calculated as the average of 200 GPU computation delays. At lower resolutions, the latencies for different methods are comparable; however, when the resolution reaches 400, the latencies for all methods increase, with global attention experiencing the most dramatic growth. Across all tested resolutions, Filter Enhancement maintains the lowest latency among the three methods. 

\subsection{Deficiencies and prospects}
The limitation of our model lies in the fact that for hierarchical models, when the feature map resolution is small, the filter enhancement method may not yield satisfactory results. This is because the frequency resolution of the feature map in the frequency domain is also low at such times, making it difficult to precisely distinguish between different frequency components, which affects the filtering effect. Meanwhile, low-resolution feature maps inherently lack high-frequency details, and information may be further lost after filtering, leading to subpar performance.

However, precisely because of this, we are hopeful about the performance of this method on high-resolution images, as they are likely to avoid several of the aforementioned issues. However, due to computational power limitations, we have not been able to conduct experiments at higher resolutions.

\section{Conclusion}

The proposed FwNet-ECA achieves information exchange between different windows in window attention through Filter Enhancement operations based on Fourier Transform, reducing the computational cost of inter-window communication from \(O(N^{2})\) to \(O(Nlog(N))\). This approach enables the model to achieve the largest receptive field with reduced parameters and computational overhead. Additionally, it establishes communication between channels through ECA modules. Experimental results confirm the effectiveness of the Filter Enhancement layer and ECA modules, and through visualization methods, it is proved that the filtering enhancement can indeed obtain a better receptive field than the shifted window attention. Furthermore, our model demonstrates certain advantages in fine-grained classification tasks. 

\section*{Declarations}

\begin{itemize}
\item Funding: This work was supported by the Capital University of Economics and Business (No. XRZ2022065). The funders had no role in study design, data collection and analysis, decision to publish, or preparation of the manuscript.
\item Conflict of interest: The authors have no competing interests to declare that are relevant to the content of this article.
\item Ethics approval: Not applicable.
\item Consent for publication: Not applicable.
\item Data availability: The datasets that support the findings of this study are openly accessible through the following resources. iCartoonFace: \href{https://github.com/luxiangju-PersonAI/iCartoonFace?tab=readme-ov-file\#Dataset}{GitHub - luxiangju-PersonAI/iCartoonFace: iCartoonFace dataset, and baseline approaches, the project is supported by iQIYI}.
cifar100: \href{https://pytorch.org/vision/stable/generated/torchvision.datasets.CIFAR100.html}{CIFAR100 — Torchvision 0.19 documentation (pytorch.org)}
Flowers102: \href{https://pytorch.org/vision/stable/generated/torchvision.datasets.Flowers102.html}{Flowers102 — Torchvision 0.19 documentation (pytorch.org)}
StanfordCars: \href{https://pytorch.org/vision/stable/generated/torchvision.datasets.StanfordCars.html}{StanfordCars — Torchvision 0.19 documentation (pytorch.org)}
\item Code availability: The code used or analysed during the current study is available from the corresponding author on reasonable request.
\item Author contribution: Shengtian Mian conceived and designed the experiments, performed the experiments, analyzed the data, performed the computation work, prepared figures and tables, reviewed drafts of the article, and approved the final draft. Ya Wang conceived and designed the experiments, analyzed the data, reviewed drafts of the article, and approved the final draft. Nannan Gu conceived and designed the experiments and approved the final draft. Yuping Wang reviewed drafts of the article and approved the final draft. Xiaoqing Li conceived and designed the experiments, analyzed the data, reviewed drafts of the article, and approved the final draft.
\end{itemize}

\bibliography{sn-bibliography}

\end{document}